\documentclass[sigconf]{acmart}

\usepackage{graphicx}
\usepackage{booktabs}
\usepackage{multirow}

\renewcommand\footnotetextcopyrightpermission[1]{}  
\title{From Text to Network: Constructing a Knowledge Graph of Taiwan-Based China Studies Using Generative AI} 

\author{Hsuan-Lei Shao}
\affiliation{
    \institution{Graduate Institute of Health and Biotechnology Law, Taipei Medical University}
    \country{Taiwan}
}
\email{hlshao@tmu.edu.tw}
\authornote{https://orcid.org/0000-0002-7101-5272}

\begin{abstract}
Taiwan’s China Studies (CS) has developed into a rich, interdisciplinary research field shaped by the island’s unique geopolitical position and long-standing academic engagement with Mainland China. This study responds to the growing need to systematically revisit and reorganize decades of Taiwan-based CS scholarship by proposing an AI-assisted approach that transforms unstructured academic texts into structured, interactive knowledge representations.

We apply generative AI (GAI) techniques and large language models (LLMs) to extract and standardize entity–relation triples from 1,367 peer-reviewed CS articles published between 1996 and 2019. These triples are then visualized through a lightweight D3.js-based system\cite{bostock2011d3}, forming the foundation of a domain-specific knowledge graph and vector database for the field. This infrastructure allows users to explore conceptual nodes and semantic relationships across the corpus, revealing previously uncharted intellectual trajectories, thematic clusters, and research gaps.

By decomposing textual content into graph-structured knowledge units, our system enables a paradigm shift from linear text consumption to network-based knowledge navigation. In doing so, it enhances scholarly access to CS literature while offering a scalable, data-driven alternative to traditional ontology construction. This work not only demonstrates how generative AI can augment area studies and digital humanities but also highlights its potential to support a reimagined scholarly infrastructure for regional knowledge systems.

\end{abstract}


\begin{document}

\maketitle

\section{Introduction}
Taiwan's "China Studies" has developed into a rich and multidisciplinary academic field, shaped by its distinctive geopolitical context and intellectual engagement with Mainland China. This field has been deeply influenced by Taiwan’s democratic transition and its complex identity politics. From the late 1990s to 2019, a critical period for cross-strait relations, Taiwan witnessed significant transformations in academic output, echoing broader societal and diplomatic shifts.

In the context of Taiwan, the academic field of "China Studies" (CS) has accumulated a rich corpus of scholarship over the past decades. With increasing concern about the need for "principle development" in political science studies \cite{grimmer2013text, mcpherson2001birds, card1999readings}, there has been a growing call to “text research on China Studies” as a new methodological approaches. This study responds to that call by applying digital methods—specifically text mining (TM) and machine learning (ML) techniques—to revisit and analyze the intellectual structure of CS community \cite{grimmer2013text, laver2003extracting}. 

Based on the data set that the authors' research team has built, a total of 1,367 papers published between 1996 and 2019 in Taiwan-based journals were collected, parsed, and structured to form a comprehensive database of "China Studies". This database facilitates the extraction and analysis of metadata, including authorship, journal affiliations, and institutional information.

\begin{figure}[htbp]
  \centering
  \includegraphics[width=0.9\linewidth]{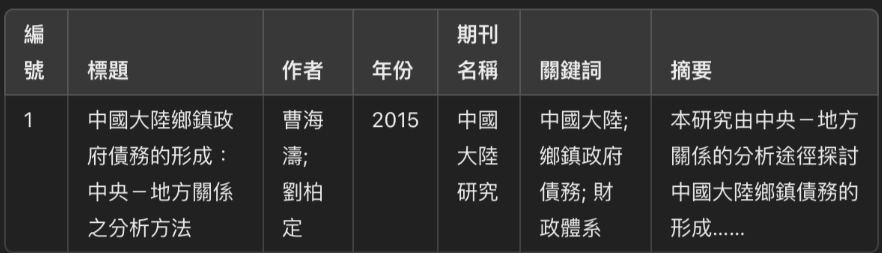}
  \caption{Dataset Structure}
  \Description{A sample knowledge graph that shows topics, authors, and institutions connected in a network.}
  \label{fig:china-studies-kg}
\end{figure}

As a further step, this study responds to that call by applying Generative AI (GAI) and Large Language Models (LLMs) to analyze and structure CS literature into a vectorized, searchable knowledge infrastructure. In parallel, we present an interactive visualization system based on a structured triplet (subject - predicate - object)\cite{bostock2011d3}, constructed to explore and interpret semantic relations in CS and political knowledge in a more general sense\cite{schutze2008introduction}. Triple-based models, widely used in semantic web and NLP applications, offer structured yet interpretable representations of entity relations. By integrating this approach with data extracted from political discourse, our visualization system bridges computational tools with domain-specific needs in political science research, particularly in tracing the shifting contours of ideological and intellectual trends within Taiwan’s scholarly landscape.

\section{Methodology}
\subsection{Triplet Annotation}

In order to build knowledge graph construction, we follow this lineage by applying triplet (Subject–Predicate–Object) and vector-based indexing techniques to map relationships among the annotations on Taiwan's China Studies papers.

The annotation process in this study was primarily supported by two large language models: GPT-4o and Breeze-7B, serving as automated triplet extraction engines. Each model independently processed the training dataset and generated candidate subject–predicate–object triples from the article metadata and abstracts. To improve the quality of extracted knowledge, we applied a filtering mechanism to remove low-value lexical items—such as generic terms like "result" or "study"—which contributed little to the semantic structure of the knowledge graph.

In cases where multiple semantically similar triplets were extracted, we adopted a selection strategy that favored clarity and thematic precision. For example, the phrase "local governments favor investing in industries" was simplified to "local governments favor investment," to ensure conceptual generalizability and structural coherence across the graph.

\begin{figure}[htbp]
  \centering
  \includegraphics[width=0.9\linewidth]{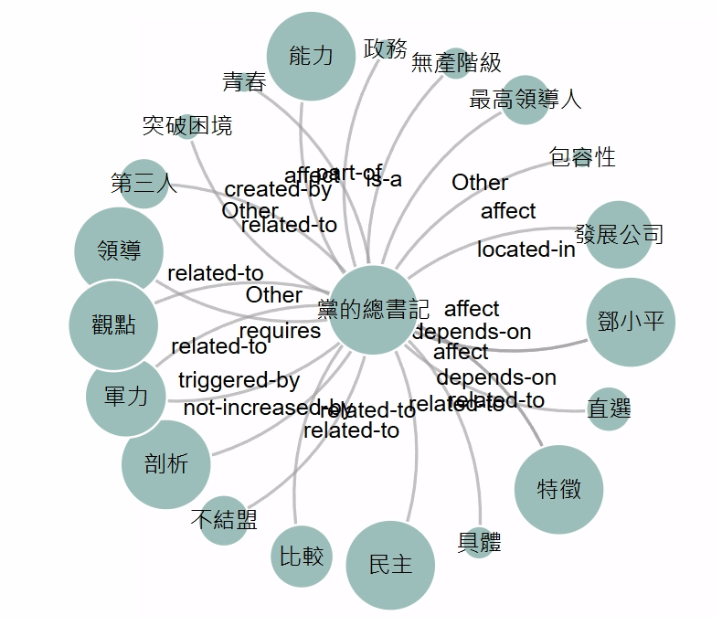}
  \caption{Basic Triplet Visualization}
  \Description{A sample knowledge graph that shows topics, authors, and institutions connected in a network.}
  \label{fig:china-studies-kg}
\end{figure}


\subsection{Preprocessing: Reduce Divergence}

The preprocessing phase is critical to ensure the consistency, clarity, and usability of relational data before visualization.\cite{benoit2018quanteda} As information visualizations, especially graph-based ones, are sensitive to redundancy and ambiguity in input data, a robust preprocessing strategy is essential. Our preprocessing pipeline consists of four essential stages, each aimed at resolving a specific type of data inconsistency and improving interpretability for downstream visual analysis.

In our raw dataset, which consists of subject-relation-object triples extracted from a wide range of sources, relation types exhibit a highly skewed distribution. While a small subset of relation types appears frequently, a long tail of categories occurs only once or twice. These infrequent categories often result from noise, inconsistent labeling, or isolated cases that are not analytically significant. Their inclusion in the visualization leads to excessive fragmentation and hinders users from identifying dominant patterns.

\begin{figure}[htbp]
  \centering
  \includegraphics[width=0.9\linewidth]{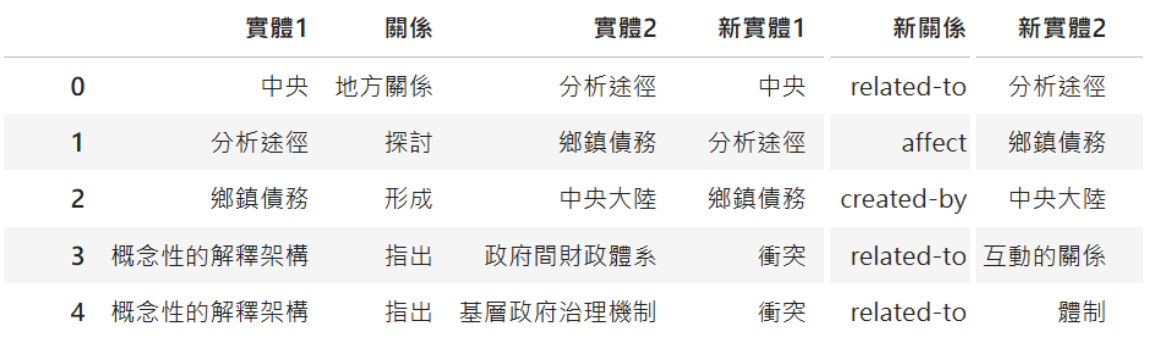}
  \caption{Original Triplet and Reduced Divergence}
  \Description{A sample knowledge graph that shows topics, authors, and institutions connected in a network.}
  \label{fig:china-studies-kg}
\end{figure}


\subsection{Preprocessing II: Semantic Label Merging}

Even after frequency-based consolidation, we observe redundancy arising from lexical variations and inconsistent terminology. For instance, relations such as \texttt{no} and \texttt{not}, or \texttt{related-to} and \texttt{related to}, essentially refer to the same underlying semantic relationship but are encoded differently due to variations in data entry or source formatting.

To resolve these inconsistencies, we apply semantic label merging. This involves grouping relation labels with similar meanings under a unified term. We implement this step manually using a semi-structured process: candidate synonyms are first identified through string similarity metrics and cross-checked against domain-specific dictionaries. We then review their contextual usage in the triple corpus to ensure that the merged terms do not conflate semantically distinct categories. This process is particularly important in knowledge graph construction, where the accuracy of relational semantics directly influences graph-based reasoning and interpretation.

\subsection{Preprocessing III: Duplicate Removal}

Redundant triples—those with identical subject, relation, and object—are a common artifact in extracted datasets, particularly when multiple data sources are aggregated. While such duplicates may reflect frequency in the original corpus, they can lead to misrepresentation when visualized as graphs. For example, repeated edges between the same pair of nodes can artificially inflate node degrees or produce overlapping labels, reducing the effectiveness of the graph layout algorithm.

We address this issue by performing an exact match deduplication pass during preprocessing. Each triple is hashed based on its constituent elements and retained only once. This ensures that the structural properties of the resulting graph, such as degree distribution and clustering coefficient, more accurately reflect distinct relationships rather than redundant mentions.

\subsection{Preprocessing IV: Abbreviation Mapping}

To enhance the readability of relation labels within the network visualization, we introduce a standardized abbreviation mapping step. Long or descriptive relation labels often exceed the space available in graph layouts, particularly when nodes are densely connected or the network is rendered at smaller scales. Abbreviations not only save space but also facilitate quicker visual parsing by experienced users.

We constructed a dedicated abbreviation lookup table that maps each full-length relation label to shorter alias (e.g., \texttt{influenced-by} $\rightarrow$ \texttt{IFB}). This table is manually curated to ensure that abbreviations are intuitive, unique, and consistent across the dataset. It is also loaded into the visualization interface to provide users with an on-demand legend for interpretation. This design balances visual economy with accessibility, supporting both novice and expert interaction with the system.


\section{Visualization System Design}

To facilitate the exploration of complex relational data, we developed an interactive network visualization system using the D3.js library, deployed in a web-based environment for maximum accessibility. The system is designed to support both high-level structural analysis and fine-grained relational inspection, enabling users to interactively explore, filter, and interpret dense networks derived from knowledge graphs or text-mined corpora. Furthermore, the system is optimized for real-time responsiveness, ensuring smooth and immediate feedback to user interactions. The key components are as follows:

\subsection{Data Loading Module}

Upon initialization, the system loads the triplet data and an abbreviation mapping table that links verbose relation names to their corresponding shortcodes.

Parsing is handled client-side via JavaScript, leveraging efficient array structures to store node and edge data in memory. During this phase, the system performs integrity checks to ensure every relation in the dataset has a corresponding abbreviation. In cases where a mismatch is detected, for example, a relation label not found in the abbreviation table, the system logs a warning to the console and flags the label for manual inspection. This quality control step is essential to ensure label consistency and avoid runtime visualization errors caused by undefined mappings.

\subsection{Layout Engine and Edge Label Management}

We adopt D3’s \texttt{forceSimulation} module as the foundation for layout generation. Each node corresponds to a unique entity (subject or object), while edges represent the labeled relations connecting them. The force-directed approach simulates physical dynamics to iteratively position nodes and edges, resulting in a graph layout that is both visually coherent and structurally informative.


Additionally, node radius is dynamically scaled according to degree centrality (i.e., the number of incoming and outgoing edges), allowing users to visually identify high-connectivity hubs that may represent key actors or concepts within the dataset.


In many real-world networks, entities may be linked via multiple distinct relations. To accurately represent such multi-edge scenarios, we implement a custom edge offset and labeling mechanism. Labels are drawn along curved SVG paths using the \texttt{path} and \texttt{textPath} elements. For each edge group connecting the same node pair, the system calculates appropriate curvature based on edge index and directionality, ensuring that labels do not stack or collide.

This dynamic label placement strategy significantly improves readability, particularly in dense subgraphs where relation differentiation is critical for interpretation. It also preserves semantic context, allowing users to distinguish between types of relationships without ambiguity.

\subsection{Interactive Filtering and Search}

The system offers multiple interactive controls to support exploratory analysis:

\begin{itemize}
    \item \textbf{Degree Threshold Slider:} Users can dynamically adjust a numeric threshold to filter out nodes below a certain degree. This enables focus on the most structurally significant entities and suppresses peripheral noise. When the threshold changes, the system recomputes the subgraph in real time and updates both the visual rendering and underlying data model.

    \item \textbf{Keyword Search and Layer Expansion:} Users can input a keyword to locate a specific entity within the graph. The system highlights the target node and optionally expands its relational neighborhood to a user-defined depth. 
    This multi-layer expansion is particularly useful for tracing the contextual embedding of a concept or actor within a broader network.\cite{huang2024comparison}
\end{itemize}

These tools are designed to support both bottom-up discovery and top-down hypothesis testing, enabling users to toggle between focused queries and global exploration seamlessly.

\subsection{User Interaction and Controls}

To enhance interactivity and facilitate in-depth exploration, the entire graph canvas supports smooth pan, zoom, and node dragging operations. Users can reposition individual nodes via click-and-drag, temporarily fixing them in place to reorganize local substructures for clearer visualization. This feature is especially useful when examining densely connected clusters or disentangling overlapping regions.

The system monitors changes in user input—such as search queries, threshold adjustments, or drag events—and responds by triggering a full simulation re-render. This ensures the visual layout remains consistent with the current interaction context while preserving spatial continuity through animated transitions.

All visual elements are rendered using Scalable Vector Graphics (SVG), which guarantees high-resolution clarity and compatibility across modern web browsers. The use of SVG also enables fine-grained styling, interactivity (e.g., tooltips on hover), and responsive layout adjustments, contributing to an overall user-friendly and visually engaging interface.

By combining real-time interactivity, semantic clarity, and flexible visual control, the system provides a powerful platform for analyzing complex relational structures—particularly in domains such as knowledge graph exploration, bibliometric mapping, or thematic clustering in political science and digital humanities.

\begin{figure}[htbp]
  \centering
  \includegraphics[width=0.9\linewidth]{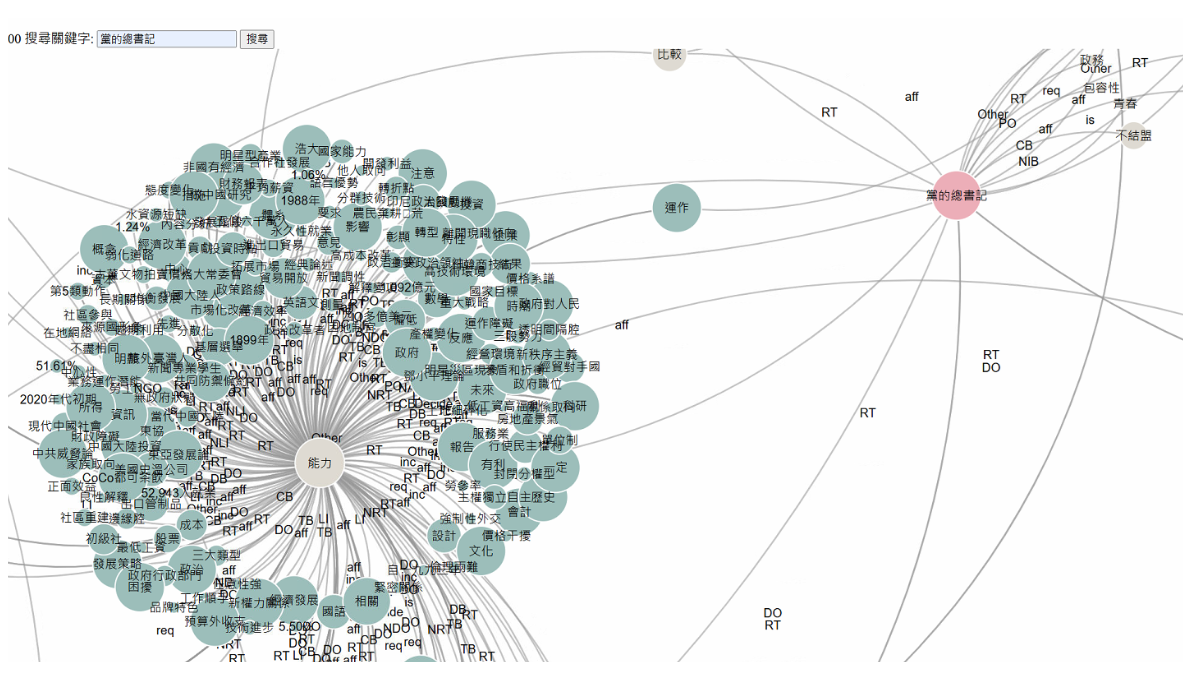}
  \caption{Knowledge Graph UI Design}
  \Description{A sample knowledge graph that shows topics, authors, and institutions connected in a network.}
  \label{fig:china-studies-kg}
\end{figure}

\section{Evaluation}
To assess the effectiveness and robustness of the proposed system, we conducted a multi-dimensional evaluation focusing on four key criteria: clarity, responsiveness, usability, and scalability. These dimensions were selected to reflect not only the visual analytic quality of the output but also the real-time interactive performance of the system in exploratory research scenarios.

Beyond conventional keyword-based retrieval systems that require extensive textual reading, our approach introduces a structurally distinct form of knowledge representation. Specifically, we extract fundamental conceptual units from each academic paper and transform them into subject–predicate–object triplets, thereby encoding each document as a semantic structure. These knowledge triplets are subsequently embedded into a larger knowledge graph, where each individual article or idea occupies a line or a region within a topological conceptual space. This allows scholars to navigate, compare, and integrate ideas across papers, enabling a more intuitive and holistic understanding of the intellectual landscape.

This representation not only enhances user experience through visual interactivity but also redefines how scholarly knowledge is accessed and expanded. New academic contributions are no longer appended as isolated bibliographic entries; instead, they are assimilated as conceptual additions to an evolving knowledge system. This conceptual granularity allows for more precise cross-paper inferences and comparative analyses.

Furthermore, our framework advances beyond traditional citation network analysis by employing large language models (LLMs) to discover latent semantic connections between research outputs. While classical ontology engineering demands extensive domain expertise and manually curated schema (e.g., OWL or RDF-based ontologies), our method adopts a lightweight, data-driven alternative. Pre-trained LLMs enable automatic semantic linking across documents without requiring a pre-defined ontology, thus significantly reducing the cost and complexity of knowledge graph construction.

This dual innovation—(1) decomposing scholarly texts into machine-readable conceptual units and (2) generating emergent knowledge structures through LLM-guided inference—demonstrates the feasibility of building semi-automated knowledge infrastructures. It not only lowers the barrier to constructing domain-specific knowledge systems but also aligns with

\section{Conclusion and Future Work}

This study presents an end-to-end system for preprocessing and visualizing triple-based knowledge graph data, with a focus on semantic clarity, user interaction, and real-time responsiveness. Through systematic data refinement—including category consolidation, label merging, duplicate removal, and abbreviation mapping—we produced a structured and interpretable dataset suitable for interactive visualization. The resulting system, implemented with D3.js and SVG-based rendering, enables exploratory analysis of complex relational structures and supports multiple entry points for user engagement.

Our work contributes to the growing body of research at the intersection of information visualization, knowledge representation, and digital humanities. In particular, we demonstrate how preprocessing quality directly impacts visualization clarity and how interactive design can enhance interpretability in knowledge graph analytics. The integration of user-centric features—such as search-layer expansion and degree-based filtering—offers a flexible environment for exploratory inquiry and hypothesis generation.

Future work will proceed along three main directions:

\begin{itemize}
    \item \textbf{User-Centered Evaluation:} We plan to conduct formal usability studies with scholars and students in political science, information studies, and digital humanities. This will provide insights into task-specific performance and user satisfaction, and help identify gaps in system functionality.

    \item \textbf{Integration with LLM-Augmented Interfaces:} Given the rise of large language models (LLMs) in natural language understanding, we aim to explore how such models can assist in automatic relation classification, semantic clustering, and even conversational querying of the graph.

    \item \textbf{Knowledge Centrality and Ontology:} Building on the foundation of social network analysis, we employ the concept of centrality\cite{freeman2002centrality, mcpherson2001birds} to identify the most influential knowledge nodes within the field of China Studies. This approach not only highlights the structural importance of specific themes and entities, even domain-specific ontology.
\end{itemize}

This research contributes directly to the core mission of the 5th World Congress of Taiwan Studies by offering an innovative digital approach to understanding the intellectual landscape of Taiwan’s academic engagement with China. By systematically mapping and visualizing the evolution of Taiwan-based "China Studies," the project facilitates new modes of scholarly access to decades of interdisciplinary research from Taiwanese.

The research team will provide a live demonstration of the web-based system and its associated findings during the conference presentation. We welcome critical feedback and constructive discussion from participants.


\bibliographystyle{ACM-Reference-Format}
\nocite{*}
\bibliography{references}

\end{document}